\documentclass[sigconf]{acmart}

\renewcommand\footnotetextcopyrightpermission[1]{}
\acmConference{}{}{}
\acmBooktitle{Preprint}

\usepackage{booktabs}
\usepackage{graphicx}
\usepackage{amsmath}

\begin{document}

\title{The Tokenizer Tax: Quantifying and Explaining the Cross-Lingual Cost of Subword Tokenization for Indian Languages}

\author{Priyansh Srivastava}
\affiliation{\institution{Sirena Ai}\country{India}}
\email{priyansh@sirenatech.com}

\begin{abstract}
Large language models do not read text; they read tokens produced by a subword tokenizer fit predominantly to English data. We show that this design choice imposes a large, systematic, and measurable \emph{tokenizer tax} on Indian languages. Using the 997-sentence FLORES-200 development set as a content-controlled parallel corpus, we measure the fertility of six widely used tokenizers across ten Indian languages and four comparison languages. Under \texttt{cl100k\_base} (the tokenizer of GPT-3.5/GPT-4), Indian languages incur a mean tax of $8.0\times$ relative to English, rising to $13.0\times$ for Malayalam---meaning an Indian-language user receives as little as $12\%$ of the effective context window an English user receives for the same content. We trace this tax to a single mechanism: the rate at which the tokenizer's byte-pair merges fail, leaving tokens stranded as unmerged single bytes, which alone correlates with the per-language tax at $r{=}0.89$. We show the tax is a remediable design choice rather than a property of the scripts: multilingual tokenizers (XLM-R) and OpenAI's newer \texttt{o200k\_base} reduce the mean Indic tax by $73\%$. We further demonstrate a model-free harm: at a fixed context budget, Indian languages retain a small fraction of real document content compared to English. Finally, we relate fertility to downstream reading-comprehension accuracy on the Belebele benchmark and find that the raw correlation is confounded by resource level, a finding we report transparently rather than overstating.
\end{abstract}


\maketitle


\section{Introduction}
A large language model (LLM) never operates on characters or words directly. Every input is first segmented by a \emph{tokenizer} into subword units drawn from a fixed vocabulary, and a model's context window, latency, and per-call cost are all denominated in these tokens~\cite{sennrich2016bpe,kudo2018subword}. Because the vocabularies of today's dominant tokenizers are learned from corpora that are overwhelmingly English, they encode English with remarkable efficiency---often close to one token per word---while shattering other scripts into many small pieces.

The number of tokens a tokenizer needs to encode a fixed unit of text is its \emph{fertility}. When fertility is high for a language, the consequences compound. The same paragraph consumes more of a fixed context window, so a Hindi or Tamil user effectively receives a \emph{smaller} window than an English user for identical content. Commercial APIs price per token, so the same meaning costs more to send and receive. Generation cost scales with token count, so responses are slower. And over-fragmentation may destroy morphological structure in ways that harm comprehension. This disparity, which we call the \emph{tokenizer tax}, was first systematically documented across many languages by Petrov et al.~\cite{petrov2023unfairness} and Ahia et al.~\cite{ahia2023cost}; we extend their measurement to explain its cause and test its remediability specifically for Indian languages.

This paper asks three questions about Indian languages specifically, which together span over a billion speakers across typologically diverse scripts:
\begin{enumerate}
\item \textbf{How large is the tax, and is it structured?} We find a clean typological gradient, not a uniform penalty.
\item \textbf{Why does it exist?} We identify a failure of BPE merging---tokens left stranded as unmerged single bytes---as the dominant, quantifiable mechanism.
\item \textbf{Is it remediable?} We show that tokenizer choice---not the scripts---drives the tax, and that recent tokenizers already reduce it sharply.
\end{enumerate}

Our contributions are: (1) a fine-grained, content-controlled measurement of the tokenizer tax across ten Indian languages and six tokenizers; (2) a mechanistic explanation via the rate of unmerged single-byte tokens, which correlates with the tax at $r{=}0.89$; (3) a cross-tokenizer analysis demonstrating that the tax is a design choice, with multilingual and newer tokenizers cutting it by $73\%$; (4) a model-free demonstration of effective-context loss under a fixed token budget; and (5) an honest analysis of fertility versus downstream accuracy that surfaces a resource-level confound rather than overclaiming causality. The entire study runs on a CPU in minutes and is fully reproducible.

\section{Related Work}
\label{sec:related}
\textbf{Subword tokenization.} Byte-Pair Encoding~\cite{sennrich2016bpe} and the unigram language-model tokenizer~\cite{kudo2018subword}, both popularized through SentencePiece~\cite{kudo2018sentencepiece}, underlie virtually all modern LLMs. Their vocabularies are optimized for compression on the training corpus, which is the root cause of the cross-lingual disparities studied here.

\textbf{Tokenization disparities across languages.} The closest prior work establishes that tokenization is unequal across languages. Petrov et al.~\cite{petrov2023unfairness} show that tokenizers introduce systematic unfairness, with some languages requiring many times more tokens than English, connecting this to context-length and cost disparities. Ahia et al.~\cite{ahia2023cost} quantify the monetary consequences of this disparity for commercial APIs. We do not claim to be first to observe that tokenizers disadvantage non-English languages; that finding is established. Our contribution is narrower and more specific: we resolve the disparity \emph{within} the Indian-language family at a finer typological grain than prior broad surveys, we identify a concrete, measurable mechanism (the rate of unmerged single-byte tokens) rather than reporting fertility ratios alone, and we show empirically how much of the disparity newer and multilingual tokenizers have already closed. We view this as a focused case study and mechanistic explanation building on Petrov et al. and Ahia et al., not a new category of finding.

\textbf{Fertility and downstream performance.} Rust et al.~\cite{rust2021tokenizer} demonstrate that a dedicated tokenizer improves a multilingual model's monolingual performance, implicating tokenizer quality in downstream accuracy. We build on this by directly correlating per-language fertility with benchmark accuracy and controlling for resource level, and we report transparently when this confound dominates the raw correlation.

\textbf{Indic NLP resources.} Multilingual models such as mBERT~\cite{devlin2019bert}, XLM-R~\cite{conneau2020xlmr}, InfoXLM~\cite{chi2021infoxlm}, and XLM-V~\cite{liang2023xlmv}, and Indic-focused efforts including IndicNLPSuite~\cite{kakwani2020indicnlp} and IndicTrans2~\cite{gala2023indictrans2}, have expanded coverage of Indian languages. We use several of their tokenizers as comparison points, and we use the Belebele benchmark~\cite{bandarkar2024belebele} for our downstream-accuracy analysis.

\section{Methodology}
\label{sec:method}
\textbf{Corpus.} We use the FLORES-200 development set~\cite{nllb2022}, which provides 997 sentences professionally translated and aligned across 200+ languages. Because every language expresses the \emph{same} content, fertility differences reflect the tokenizer and writing system rather than differences in what is being said.

\textbf{Languages.} We study ten Indian languages spanning two families and five scripts: Hindi, Marathi (Devanagari); Bengali (Bengali); Punjabi (Gurmukhi); Gujarati (Gujarati); Urdu (Perso-Arabic); and the four major Dravidian languages Tamil, Telugu, Kannada, and Malayalam, each in its own script. As comparison points we include English, Spanish, French, and Arabic.

\textbf{Tokenizers.} We evaluate six tokenizers spanning vendors and generations: OpenAI's \texttt{cl100k\_base} (GPT-3.5/GPT-4) and \texttt{o200k\_base} (GPT-4o); GPT-2~\cite{radford2019gpt2}; Qwen2.5; and the multilingual mBERT~\cite{devlin2019bert} and XLM-R~\cite{conneau2020xlmr} tokenizers.

\textbf{Metrics.} For each (tokenizer, language) pair we report: \emph{word fertility} (tokens per whitespace-delimited word); \emph{character fertility} (tokens per character); \emph{bytes per token} (encoding efficiency); and \emph{unmerged single-byte rate}\footnote{We avoid the term ``byte fallback'' in its strict sense, since \texttt{cl100k\_base} is a byte-level BPE tokenizer for \emph{every} language: all tokens are ultimately built from bytes, including English ones. What we measure is the rate at which the BPE merge process \emph{failed} to combine adjacent bytes into a larger learned subword, leaving them stranded as single-byte tokens---the practical symptom of insufficient vocabulary coverage for a script.}, the fraction of emitted tokens that surface as a single raw byte or the Unicode replacement character rather than a multi-byte learned subword. The headline quantity is the \emph{tax multiplier}: the ratio of a language's word fertility to English's under the same tokenizer. For a fixed token budget $B$, we compute usable characters as $B/(\text{char fertility})$ and its ratio to English's, the \emph{context shrinkage}.

\section{The Tax and Its Structure}
\label{sec:results}
\begin{figure}[t]
\centering
\includegraphics[width=\columnwidth]{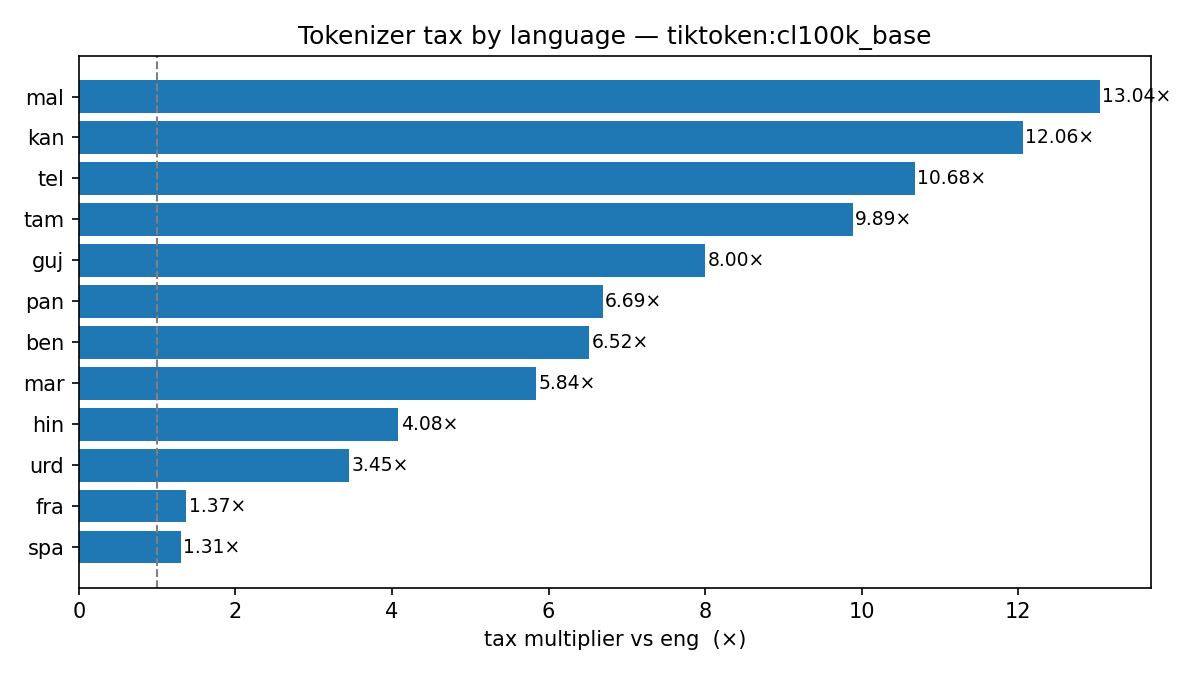}
\caption{Tokenizer tax by language under \texttt{cl100k\_base}, relative to English. The penalty follows a typological gradient, heaviest for the Dravidian languages.}
\label{fig:tax}
\end{figure}

Table~\ref{tab:main} reports measurements under \texttt{cl100k\_base}. The tax is large and \emph{structured}: it is not a flat penalty on ``non-English'' but a smooth typological gradient (Figure~\ref{fig:tax}). Latin-script European languages pay only $1.3$--$1.6\times$. Perso-Arabic scripts (Urdu, Arabic) sit near $3.4\times$. Indo-Aryan Brahmi scripts climb from Hindi ($4.1\times$) through Bengali ($6.5\times$) and Gujarati ($8.0\times$), and the Dravidian languages are taxed most heavily: Tamil ($9.9\times$), Telugu ($10.7\times$), Kannada ($12.1\times$), and Malayalam ($13.0\times$). Across the ten Indian languages the mean tax is $8.0\times$.

\begin{figure}[t]
\centering
\includegraphics[width=\columnwidth]{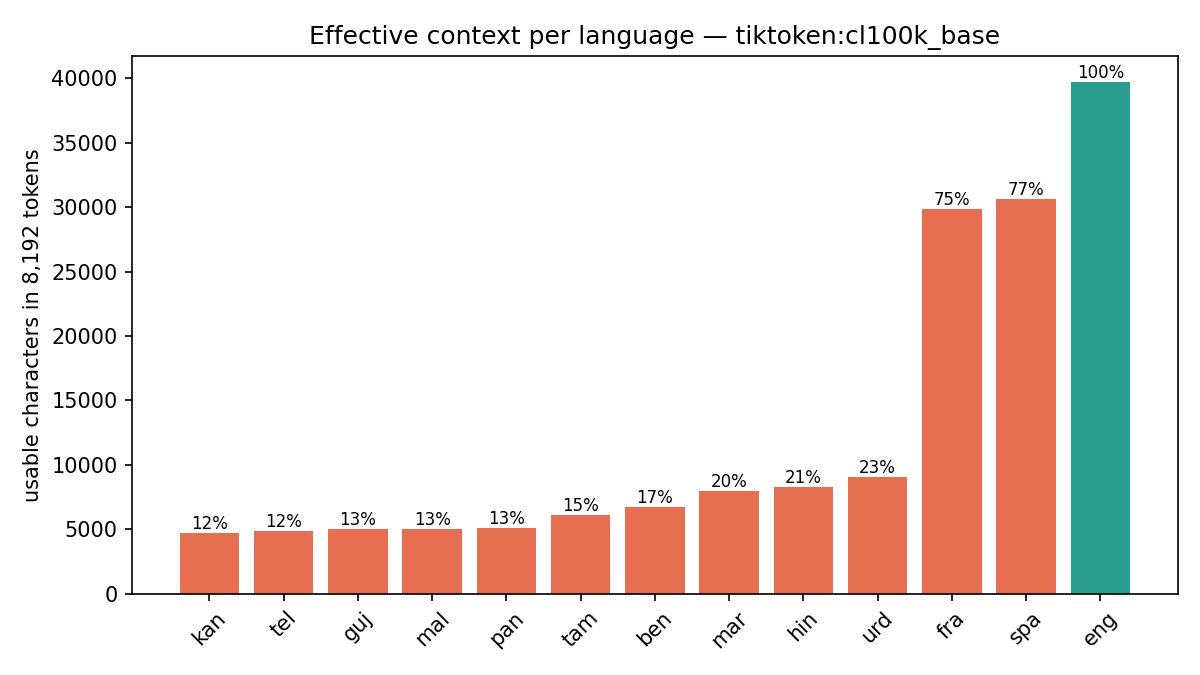}
\caption{Effective context per language: usable characters within an 8{,}192-token budget. Indian languages receive $12$--$23\%$ of the English window.}
\label{fig:context}
\end{figure}

The effective-context consequence is severe (Figure~\ref{fig:context}). At a fixed 8{,}192-token budget, Kannada and Telugu users receive only $\sim$$12\%$ of the usable characters an English user does; every Indian language studied receives under a quarter.

\begin{table}[t]
\centering
\small
\caption{Per-language measurements under \texttt{cl100k\_base}. Tax = word-fertility ratio vs.\ English; Byte/tok = bytes per token; Unmerged = unmerged single-byte token rate; Ctx = usable characters in an 8{,}192-token budget relative to English.}
\label{tab:main}
\begin{tabular}{lrrrr}
\toprule
Lang & Tax $\times$ & Byte/tok & Unmerged \% & Ctx \% \\
\midrule
eng & 1.00 & 4.86 & 9.3 & 100.0 \\
\midrule
spa & 1.31 & 3.81 & 7.1 & 77.1 \\
fra & 1.37 & 3.78 & 7.2 & 75.1 \\
\midrule
urd & 3.45 & 1.96 & 13.3 & 22.7 \\
hin & 4.08 & 2.61 & 27.5 & 20.9 \\
mar & 5.84 & 2.58 & 29.2 & 19.9 \\
ben & 6.52 & 2.19 & 37.2 & 16.9 \\
pan & 6.69 & 1.61 & 39.1 & 12.9 \\
guj & 8.00 & 1.59 & 39.5 & 12.5 \\
tam & 9.89 & 2.03 & 33.9 & 15.3 \\
tel & 10.68 & 1.57 & 40.9 & 12.2 \\
kan & 12.06 & 1.56 & 41.0 & 11.9 \\
mal & 13.04 & 1.70 & 42.7 & 12.7 \\
\midrule
arb & 3.39 & 2.56 & 4.1 & 29.1 \\
\bottomrule
\end{tabular}
\end{table}

\section{Why the Tax Exists: Unmerged Single-Byte Tokens}
\label{sec:mechanism}
When a tokenizer's learned vocabulary contains no subword covering a span of text---common for scripts under-represented in training---its BPE merges fail to combine that span's bytes into larger units, so it decomposes into individual single-byte tokens and fertility approaches the byte length of the text. Table~\ref{tab:main}'s rightmost numeric column tells the story: English and European languages emit unmerged single-byte tokens for under $10\%$ of their tokens, whereas the high-tax Indic languages do so for $27$--$43\%$. Across the languages with valid word boundaries, this single-byte rate alone correlates with the tax multiplier at \textbf{$r{=}0.89$}. The unmerged single-byte rate explains most of the observed variance in tokenizer tax, suggesting that insufficient vocabulary coverage is a dominant mechanism, made concrete at the level of individual tokens. This also explains why Urdu and Arabic, despite sharing the Perso-Arabic script, sit at a comparatively low $\sim$$3.4\times$ tax: that script is sufficiently represented in \texttt{cl100k\_base}'s training data that its BPE merges succeed and the single-byte rate stays low, whereas the Dravidian scripts are sparse enough in the training data that merges routinely fail.

\section{The Tax Is a Choice, Not a Property of the Script}
\label{sec:choice}
If the tax were inevitable for complex scripts, no tokenizer could avoid it. Table~\ref{tab:cross} shows it is not. GPT-2 taxes Malayalam at $22\times$; the multilingual XLM-R tokenizer taxes the same language at $2.0\times$, and taxes Urdu and Hindi at near parity. Multilingual training, which exposes the tokenizer to these scripts, very nearly eliminates the tax. Strikingly, OpenAI's own tokenizer improved sharply between generations: moving from \texttt{cl100k\_base} to \texttt{o200k\_base} cuts the mean Indic tax from $8.0\times$ to $2.1\times$, a \textbf{$73\%$ reduction}, with no change to the underlying scripts. The fairness of a model toward Indian languages is, to a first approximation, a decision made at tokenizer-training time.

\begin{table}[t]
\centering
\small
\setlength{\tabcolsep}{3pt}
\caption{Tax multiplier (vs.\ English) across six tokenizers.}
\label{tab:cross}
\begin{tabular}{lrrrrrr}
\toprule
Lang & cl100k & o200k & gpt2 & Qwen & mBERT & XLM-R \\
\midrule
eng & 1.00 & 1.00 & 1.00 & 1.00 & 1.00 & 1.00 \\
spa & 1.31 & 1.11 & 1.68 & 1.29 & 1.09 & 1.01 \\
urd & 3.45 & 1.31 & 4.93 & 2.50 & 1.34 & 0.98 \\
hin & 4.08 & 1.34 & 6.34 & 3.79 & 1.50 & 1.08 \\
mar & 5.84 & 2.11 & 9.04 & 5.33 & 2.29 & 1.41 \\
ben & 6.52 & 1.91 & 10.73 & 5.65 & 2.10 & 1.54 \\
pan & 6.69 & 2.23 & 6.70 & 6.19 & 1.88 & 1.34 \\
guj & 8.00 & 1.87 & 12.70 & 7.03 & 2.49 & 1.48 \\
tam & 9.89 & 2.57 & 20.04 & 7.92 & 2.74 & 1.75 \\
tel & 10.68 & 2.49 & 16.67 & 9.02 & 2.74 & 1.71 \\
kan & 12.06 & 2.72 & 18.46 & 9.42 & 3.01 & 1.86 \\
mal & 13.04 & 2.85 & 22.06 & 10.54 & 3.80 & 2.00 \\
\midrule
Gini & 0.35 & 0.19 & 0.39 & 0.33 & 0.22 & 0.14 \\
\bottomrule
\end{tabular}
\end{table}

\section{From Cost to Consequence}
\label{sec:phaseb}
\textbf{Effective-context loss is model-free harm.} The context-shrinkage result (Figure~\ref{fig:context}) demonstrates harm without invoking model accuracy. Under any fixed token budget, high-fertility languages admit a fraction of the real content. Content that fits for an English user is truncated for an Indian-language user purely as a function of tokenization---a deterministic consequence, not a statistical tendency.

\begin{figure}[t]
\centering
\includegraphics[width=\columnwidth]{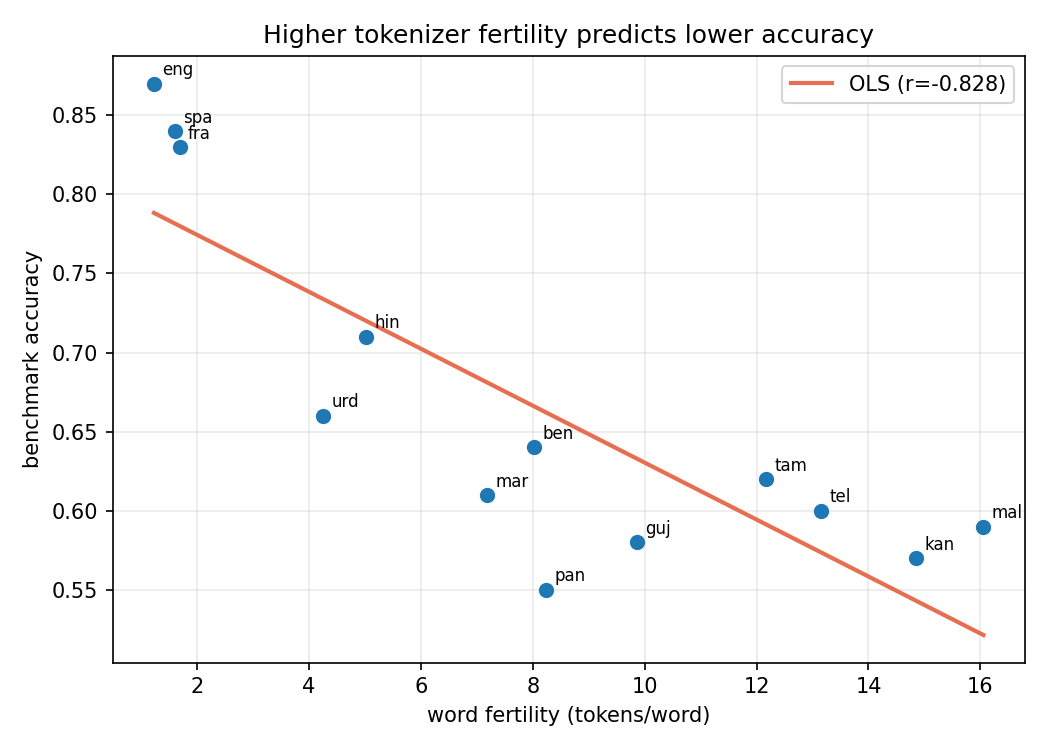}
\caption{Word fertility vs.\ Belebele accuracy (InfoXLM, Translate-Train-All). The Indian languages cluster at similar accuracy across a wide fertility range, suggesting a resource-level threshold effect rather than a smooth fertility effect.}
\label{fig:scatter}
\end{figure}

\textbf{Fertility versus downstream accuracy.} We correlate per-language word fertility under \texttt{cl100k\_base} with published per-language reading-comprehension accuracy from the Belebele benchmark~\cite{bandarkar2024belebele}, using InfoXLM's Translate-Train-All scores~\cite{chi2021infoxlm} as the accuracy signal for our thirteen languages (Arabic is excluded here, as a matching Belebele/InfoXLM score was unavailable to us), with a log-resource proxy as a covariate. The raw correlation is moderate and negative ($r=-0.61$, 95\% CI $[-0.86,-0.03]$, $n{=}13$): languages with higher fertility tend to score lower. However, the \emph{partial} correlation controlling for resource level is small and positive ($r=0.25$), indicating that in this dataset, the apparent fertility--accuracy relationship is substantially explained by resource level rather than by fertility independently. Visual inspection (Figure~\ref{fig:scatter}) clarifies why: English and the European languages cluster at high fertility-efficiency and high accuracy, while the ten Indian languages cluster together at lower accuracy across a wide range of fertility values (Hindi at $4.1\times$ scores similarly to Malayalam at $13.0\times$). This pattern is more consistent with a \emph{threshold effect}---non-Latin, lower-resource languages underperforming as a group---than with a smooth dose-response relationship between fertility and accuracy. We report this transparently: it tempers the causal interpretation of Section~\ref{sec:mechanism} and Section~\ref{sec:choice} without weakening them, since the context-loss result and the unmerged-single-byte mechanism are model-free and stand independently of this regression.

\textbf{The cost framing.} The tax is also directly financial. Because APIs bill per token, a conversation that costs an English user one unit costs a Malayalam user roughly thirteen units under \texttt{cl100k\_base} for identical content---a regressive charge falling on speakers of exactly the languages least served by the technology~\cite{ahia2023cost}.

\section{Discussion}
The tokenizer tax on Indian languages is large ($8\times$ on average, $13\times$ at worst under a frontier-model tokenizer), structured along typological lines, mechanistically driven by a failure of BPE merging that leaves Indic text fragmented into unmerged single-byte tokens, and a remediable consequence of vocabulary coverage rather than an inherent cost of complex scripts. Newer and multilingual tokenizers already demonstrate that most of the tax can be removed. The downstream-accuracy picture is more nuanced: our data support a model-free context-loss harm and a strong mechanistic explanation, but not yet an independent causal claim that fertility---separate from resource level---degrades accuracy. We view this as an honest finding rather than a weakness: future work with finer-grained resource controls or controlled tokenizer interventions is needed to disentangle these factors.

Three recommendations follow. For \emph{model builders}: tokenizer vocabulary coverage for Indic scripts is a high-leverage, low-cost fairness intervention, and the unmerged single-byte rate is a simple diagnostic to monitor. For \emph{API providers}: per-token pricing is regressive across languages. For \emph{practitioners} building Indic applications: tokenizer choice materially changes effective context and cost, and should be selected deliberately.

\section{Limitations}
Word fertility depends on whitespace word segmentation; we therefore rely on character fertility for scripts without word spaces (not used as headline languages here). FLORES sentences, while content-controlled, are translations and may exhibit translationese. Detection of unmerged single-byte tokens is best-effort and tokenizer-specific. The accuracy data in Section~\ref{sec:phaseb} comes from a different tokenizer (InfoXLM's) than the fertility measurements (\texttt{cl100k\_base}'s), since GPT-3.5/4 per-language accuracy figures were not available to us in machine-readable form; this is a limitation we flag explicitly, and the $n{=}13$ sample is modest. Establishing causality for the fertility--accuracy relationship would require controlled interventions on the tokenizer, which we leave to future work.

\section{Conclusion}
We have quantified the tokenizer tax on Indian languages, explained it through a failure of BPE merging that leaves Indic text fragmented into unmerged single-byte tokens, shown it to be a remediable design choice that newer tokenizers already reduce by $73\%$, and demonstrated a model-free context-loss harm. Our downstream-accuracy analysis is reported transparently, including a confound that tempers (without negating) the broader finding. The study requires no GPU and is fully reproducible. As LLMs become infrastructure for a billion Indian-language speakers, the tokenizer deserves attention as a first-order determinant of equity.

\section*{Ethics and Privacy Statement}
This work analyzes publicly available tokenizers and the FLORES-200 and Belebele benchmarks; it does not collect, process, or release any personal or sensitive data. The societal aim of this paper is corrective: by quantifying a fairness gap that disadvantages Indian-language speakers in current LLM tokenization, we hope to motivate vocabulary-coverage improvements rather than to disadvantage any language or community. We see no plausible misuse of these findings beyond their stated purpose of measuring and explaining an existing disparity.

\section*{Conflict of Interest and Funding}
The authors declare no conflict of interest. This research received no external funding.

\section*{Reproducibility}
The code, measurement pipeline, and scripts used in this study will be publicly released after the peer-review process. The arXiv version of this paper will be updated with a link to the repository upon release.

\bibliographystyle{ACM-Reference-Format}
\bibliography{References}


\begin{thebibliography}{15}


\ifx \showCODEN    \undefined \def \showCODEN     #1{\unskip}     \fi
\ifx \showISBNx    \undefined \def \showISBNx     #1{\unskip}     \fi
\ifx \showISBNxiii \undefined \def \showISBNxiii  #1{\unskip}     \fi
\ifx \showISSN     \undefined \def \showISSN      #1{\unskip}     \fi
\ifx \showLCCN     \undefined \def \showLCCN      #1{\unskip}     \fi
\ifx \shownote     \undefined \def \shownote      #1{#1}          \fi
\ifx \showarticletitle \undefined \def \showarticletitle #1{#1}   \fi
\ifx \showURL      \undefined \def \showURL       {\relax}        \fi
\providecommand\bibfield[2]{#2}
\providecommand\bibinfo[2]{#2}
\providecommand\natexlab[1]{#1}
\providecommand\showeprint[2][]{arXiv:#2}

\bibitem[Ahia et~al\mbox{.}(2023)]%
        {ahia2023cost}
\bibfield{author}{\bibinfo{person}{Orevaoghene Ahia}, \bibinfo{person}{Sachin Kumar}, \bibinfo{person}{Hila Gonen}, \bibinfo{person}{Jungo Kasai}, \bibinfo{person}{David~R. Mortensen}, \bibinfo{person}{Noah~A. Smith}, {and} \bibinfo{person}{Yulia Tsvetkov}.} \bibinfo{year}{2023}\natexlab{}.
\newblock \showarticletitle{Do All Languages Cost the Same? Tokenization in the Era of Commercial Language Models}. In \bibinfo{booktitle}{\emph{Proceedings of the 2023 Conference on Empirical Methods in Natural Language Processing (EMNLP)}}. \bibinfo{pages}{9904--9923}.
\newblock


\bibitem[Bandarkar et~al\mbox{.}(2024)]%
        {bandarkar2024belebele}
\bibfield{author}{\bibinfo{person}{Lucas Bandarkar}, \bibinfo{person}{Davis Liang}, \bibinfo{person}{Benjamin Muller}, \bibinfo{person}{Mikel Artetxe}, \bibinfo{person}{Satya~Narayan Shukla}, \bibinfo{person}{Donald Husa}, \bibinfo{person}{Naman Goyal}, \bibinfo{person}{Abhinandan Krishnan}, \bibinfo{person}{Luke Zettlemoyer}, {and} \bibinfo{person}{Madian Khabsa}.} \bibinfo{year}{2024}\natexlab{}.
\newblock \showarticletitle{The Belebele Benchmark: A Parallel Reading Comprehension Dataset in 122 Language Variants}. In \bibinfo{booktitle}{\emph{Proceedings of the 62nd Annual Meeting of the Association for Computational Linguistics (Volume 1: Long Papers)}}. \bibinfo{pages}{749--775}.
\newblock


\bibitem[Chi et~al\mbox{.}(2021)]%
        {chi2021infoxlm}
\bibfield{author}{\bibinfo{person}{Zewen Chi}, \bibinfo{person}{Li Dong}, \bibinfo{person}{Furu Wei}, \bibinfo{person}{Nan Yang}, \bibinfo{person}{Saksham Singhal}, \bibinfo{person}{Wenhui Wang}, \bibinfo{person}{Xia Song}, \bibinfo{person}{Xian-Ling Mao}, \bibinfo{person}{Heyan Huang}, {and} \bibinfo{person}{Ming Zhou}.} \bibinfo{year}{2021}\natexlab{}.
\newblock \showarticletitle{InfoXLM: An Information-Theoretic Framework for Cross-Lingual Language Model Pre-Training}. In \bibinfo{booktitle}{\emph{Proceedings of the 2021 Conference of the North American Chapter of the Association for Computational Linguistics: Human Language Technologies (NAACL-HLT)}}. \bibinfo{pages}{3576--3588}.
\newblock


\bibitem[Conneau et~al\mbox{.}(2020)]%
        {conneau2020xlmr}
\bibfield{author}{\bibinfo{person}{Alexis Conneau}, \bibinfo{person}{Kartikay Khandelwal}, \bibinfo{person}{Naman Goyal}, \bibinfo{person}{Vishrav Chaudhary}, \bibinfo{person}{Guillaume Wenzek}, \bibinfo{person}{Francisco Guzm{\'a}n}, \bibinfo{person}{Edouard Grave}, \bibinfo{person}{Myle Ott}, \bibinfo{person}{Luke Zettlemoyer}, {and} \bibinfo{person}{Veselin Stoyanov}.} \bibinfo{year}{2020}\natexlab{}.
\newblock \showarticletitle{Unsupervised Cross-lingual Representation Learning at Scale}. In \bibinfo{booktitle}{\emph{Proceedings of the 58th Annual Meeting of the Association for Computational Linguistics (ACL)}}. \bibinfo{pages}{8440--8451}.
\newblock


\bibitem[Devlin et~al\mbox{.}(2019)]%
        {devlin2019bert}
\bibfield{author}{\bibinfo{person}{Jacob Devlin}, \bibinfo{person}{Ming-Wei Chang}, \bibinfo{person}{Kenton Lee}, {and} \bibinfo{person}{Kristina Toutanova}.} \bibinfo{year}{2019}\natexlab{}.
\newblock \showarticletitle{BERT: Pre-training of Deep Bidirectional Transformers for Language Understanding}. In \bibinfo{booktitle}{\emph{Proceedings of the 2019 Conference of the North American Chapter of the Association for Computational Linguistics (NAACL-HLT)}}. \bibinfo{pages}{4171--4186}.
\newblock


\bibitem[Gala et~al\mbox{.}(2023)]%
        {gala2023indictrans2}
\bibfield{author}{\bibinfo{person}{Jay Gala}, \bibinfo{person}{Pranjal~A. Chitale}, \bibinfo{person}{A.~K. Raghavan}, \bibinfo{person}{Varun Gumma}, \bibinfo{person}{Sumanth Doddapaneni}, {et~al\mbox{.}}} \bibinfo{year}{2023}\natexlab{}.
\newblock \showarticletitle{IndicTrans2: Towards High-Quality and Accessible Machine Translation Models for all 22 Scheduled Indian Languages}.
\newblock \bibinfo{journal}{\emph{Transactions on Machine Learning Research (TMLR)}} (\bibinfo{year}{2023}).
\newblock


\bibitem[Kakwani et~al\mbox{.}(2020)]%
        {kakwani2020indicnlp}
\bibfield{author}{\bibinfo{person}{Divyanshu Kakwani}, \bibinfo{person}{Anoop Kunchukuttan}, \bibinfo{person}{Satish Golla}, \bibinfo{person}{Gokul N.C.}, \bibinfo{person}{Avik Bhattacharyya}, \bibinfo{person}{Mitesh~M. Khapra}, {and} \bibinfo{person}{Pratyush Kumar}.} \bibinfo{year}{2020}\natexlab{}.
\newblock \showarticletitle{IndicNLPSuite: Monolingual Corpora, Evaluation Benchmarks and Pre-trained Multilingual Language Models for Indian Languages}. In \bibinfo{booktitle}{\emph{Findings of the Association for Computational Linguistics: EMNLP 2020}}. \bibinfo{pages}{4948--4961}.
\newblock


\bibitem[Kudo(2018)]%
        {kudo2018subword}
\bibfield{author}{\bibinfo{person}{Taku Kudo}.} \bibinfo{year}{2018}\natexlab{}.
\newblock \showarticletitle{Subword Regularization: Improving Neural Network Translation Models with Multiple Subword Candidates}. In \bibinfo{booktitle}{\emph{Proceedings of the 56th Annual Meeting of the Association for Computational Linguistics (ACL)}}. \bibinfo{pages}{66--75}.
\newblock


\bibitem[Kudo and Richardson(2018)]%
        {kudo2018sentencepiece}
\bibfield{author}{\bibinfo{person}{Taku Kudo} {and} \bibinfo{person}{John Richardson}.} \bibinfo{year}{2018}\natexlab{}.
\newblock \showarticletitle{SentencePiece: A Simple and Language Independent Subword Tokenizer and Detokenizer for Neural Text Processing}. In \bibinfo{booktitle}{\emph{Proceedings of the 2018 Conference on Empirical Methods in Natural Language Processing (EMNLP): System Demonstrations}}. \bibinfo{pages}{66--71}.
\newblock


\bibitem[Liang et~al\mbox{.}(2023)]%
        {liang2023xlmv}
\bibfield{author}{\bibinfo{person}{Davis Liang}, \bibinfo{person}{Hila Gonen}, \bibinfo{person}{Yuning Mao}, \bibinfo{person}{Rui Hou}, \bibinfo{person}{Naman Goyal}, \bibinfo{person}{Marjan Ghazvininejad}, \bibinfo{person}{Luke Zettlemoyer}, {and} \bibinfo{person}{Madian Khabsa}.} \bibinfo{year}{2023}\natexlab{}.
\newblock \showarticletitle{XLM-V: Overcoming the Vocabulary Bottleneck in Multilingual Masked Language Models}. In \bibinfo{booktitle}{\emph{Proceedings of the 2023 Conference on Empirical Methods in Natural Language Processing (EMNLP)}}. \bibinfo{pages}{13142--13152}.
\newblock


\bibitem[{NLLB Team}(2022)]%
        {nllb2022}
\bibfield{author}{\bibinfo{person}{{NLLB Team}}.} \bibinfo{year}{2022}\natexlab{}.
\newblock \showarticletitle{No Language Left Behind: Scaling Human-Centered Machine Translation}.
\newblock \bibinfo{journal}{\emph{arXiv preprint arXiv:2207.04672}} (\bibinfo{year}{2022}).
\newblock


\bibitem[Petrov et~al\mbox{.}(2023)]%
        {petrov2023unfairness}
\bibfield{author}{\bibinfo{person}{Aleksandar Petrov}, \bibinfo{person}{Emanuele La~Malfa}, \bibinfo{person}{Philip H.~S. Torr}, {and} \bibinfo{person}{Adel Bibi}.} \bibinfo{year}{2023}\natexlab{}.
\newblock \showarticletitle{Language Model Tokenizers Introduce Unfairness Between Languages}. In \bibinfo{booktitle}{\emph{Advances in Neural Information Processing Systems (NeurIPS)}}.
\newblock


\bibitem[Radford et~al\mbox{.}(2019)]%
        {radford2019gpt2}
\bibfield{author}{\bibinfo{person}{Alec Radford}, \bibinfo{person}{Jeffrey Wu}, \bibinfo{person}{Rewon Child}, \bibinfo{person}{David Luan}, \bibinfo{person}{Dario Amodei}, {and} \bibinfo{person}{Ilya Sutskever}.} \bibinfo{year}{2019}\natexlab{}.
\newblock \bibinfo{booktitle}{\emph{Language Models are Unsupervised Multitask Learners}}.
\newblock \bibinfo{type}{{T}echnical {R}eport}. \bibinfo{institution}{OpenAI}.
\newblock


\bibitem[Rust et~al\mbox{.}(2021)]%
        {rust2021tokenizer}
\bibfield{author}{\bibinfo{person}{Phillip Rust}, \bibinfo{person}{Jonas Pfeiffer}, \bibinfo{person}{Ivan Vuli{\'c}}, \bibinfo{person}{Sebastian Ruder}, {and} \bibinfo{person}{Iryna Gurevych}.} \bibinfo{year}{2021}\natexlab{}.
\newblock \showarticletitle{How Good is Your Tokenizer? On the Monolingual Performance of Multilingual Language Models}. In \bibinfo{booktitle}{\emph{Proceedings of the 59th Annual Meeting of the Association for Computational Linguistics (ACL)}}. \bibinfo{pages}{3118--3135}.
\newblock


\bibitem[Sennrich et~al\mbox{.}(2016)]%
        {sennrich2016bpe}
\bibfield{author}{\bibinfo{person}{Rico Sennrich}, \bibinfo{person}{Barry Haddow}, {and} \bibinfo{person}{Alexandra Birch}.} \bibinfo{year}{2016}\natexlab{}.
\newblock \showarticletitle{Neural Machine Translation of Rare Words with Subword Units}. In \bibinfo{booktitle}{\emph{Proceedings of the 54th Annual Meeting of the Association for Computational Linguistics (ACL)}}. \bibinfo{pages}{1715--1725}.
\newblock


\end{thebibliography}

\end{document}